\documentclass[letterpaper]{article} 
\usepackage{aaai2026}  
\usepackage{times}  
\usepackage{helvet}  
\usepackage{courier}  
\usepackage[hyphens]{url}  
\usepackage{graphicx} 
\usepackage{natbib}  
\usepackage{caption} 
\usepackage{algorithm}
\usepackage{algpseudocode}
\usepackage{booktabs}
\usepackage{amssymb}
\usepackage{bm}
\usepackage{enumitem}
\usepackage{mathtools}
\usepackage{gensymb}
\usepackage{textcomp}
\usepackage{marvosym}
\usepackage{newfloat}
\usepackage{listings}
\usepackage{multirow}
\usepackage{subcaption}
\usepackage{fontawesome}
\usepackage{xcolor}

\DeclareCaptionStyle{ruled}{labelfont=normalfont,labelsep=colon,strut=off} 
\floatstyle{ruled}
\newfloat{listing}{tb}{lst}{}
\floatname{listing}{Listing}
\title{Revitalizing Canonical Pre-Alignment for Irregular Multivariate \\ Time Series Forecasting}

\author{
  Ziyu Zhou\textsuperscript{1},
  Yiming Huang\textsuperscript{1},
  Yanyun Wang\textsuperscript{1},
  Yuankai Wu\textsuperscript{2},
  James Kwok\textsuperscript{3}\thanks{Corresponding authors.},
  Yuxuan Liang\textsuperscript{1}\footnotemark[1]\\
}

\affiliations{
\textsuperscript{\rm 1}The Hong Kong University of Science and Technology (Guangzhou)\\
    \textsuperscript{\rm 2}Sichuan University 
    \textsuperscript{\rm 3}The Hong Kong University of Science and Technology \\
    \{zzhou651, yhuang033, ywang856\}@connect.hkust-gz.edu.cn, \\
    wuyk0@scu.edu.cn, jamesk@cse.ust.hk, yuxliang@outlook.com
}

\begin{document}

\maketitle

\begin{abstract}
Irregular multivariate time series (IMTS), characterized by uneven sampling and inter‑variate asynchrony, fuel many forecasting applications yet remain challenging to model efficiently. Canonical Pre‑Alignment (CPA) has been widely adopted in IMTS modeling by padding zeros at every global timestamp, thereby alleviating inter-variate asynchrony and unifying the series length, but its dense zero‑padding inflates the pre‑aligned series length, especially when numerous variates are present, causing prohibitive compute overhead. Recent graph‑based models with patching strategies sidestep CPA, but their local message passing struggles to capture global inter‑variate correlations. Therefore, we posit that CPA should be retained, with the pre‑aligned series properly handled by the model, enabling it to outperform state‑of‑the‑art graph‑based baselines that sidestep CPA. Technically, we propose \textbf{KAFNet}, a compact architecture grounded in CPA for IMTS forecasting that couples (1) Pre‑Convolution module for sequence smoothing and sparsity mitigation, (2) Temporal \textbf{K}ernel \textbf{A}ggregation module for learnable compression and modeling of intra-series irregularity, and (3) \textbf{F}requency Linear Attention blocks for the low‑cost inter-series correlations modeling in the frequency domain. Experiments on multiple IMTS datasets show that KAFNet achieves state-of-the-art forecasting performance, with a \textbf{7.2$\times$} parameter reduction and a \textbf{8.4$\times$} training-inference acceleration. The source code can be accessed at \url{https://github.com/zhouziyu02/KAFNet}.

\end{abstract}

\section{Introduction}
\label{sec:intro}

Irregular time series are prevalent in various real-world scenarios, ranging from transportation \citep{ASeer} to meteorology \citep{SPECTRUM}. In modern sensing systems, sensor malfunctions, transmission errors, and cost‑driven sampling strategies commonly give rise to irregular multivariate time series (IMTS) \citep{MuSiCNet,TimeCHEAT,GenCast}, in which observations are (i) unevenly spaced within each variate (intra‑series irregularity) and (ii) mutually asynchronous across variates (inter‑series asynchrony) \citep{Unleash,tPatchGNN,GraFITi}. These twin forms of irregularity greatly complicate the modeling of long-term temporal dependency and inter-variate correlations, as most classical approaches (e.g., RNN) implicitly assume regularly spaced and synchronously aligned series (a.k.a., regular multivariate time series, MTS) \citep{TSFool,shen2021time,shen2023non, ruan,TimesURL,Dualcast,fang2025,SDformer,wavets}.

\begin{figure}[t]
    \centering
    \includegraphics[width=0.9\linewidth]{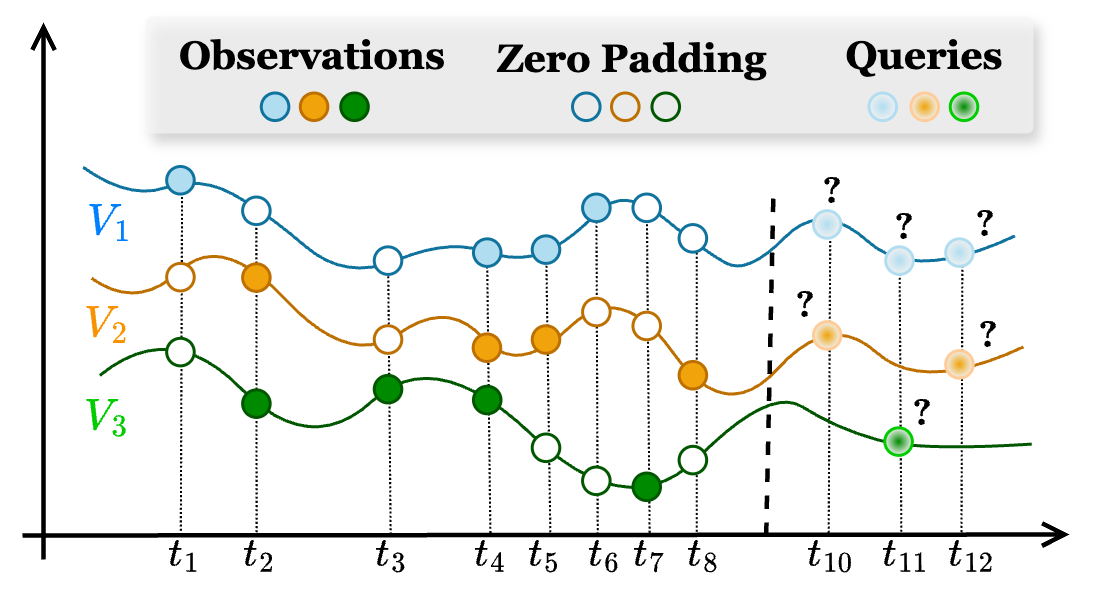}
    \caption{Illustration of Canonical Pre‑Alignment (CPA).}
    \label{fig:cpa}
\end{figure}

In order to alleviate inter‑series asynchrony, Canonical Pre‑Alignment (CPA) has become a widely adopted preprocessing procedure in IMTS modeling \citep{Che2018,Latent-ODE,mTAND,Neural-Flows,CSDI,RainDrop,Warpformer}. As illustrated in Fig.~\ref{fig:cpa}, CPA aligns all variates onto a shared temporal grid by inserting zero‑valued placeholders at every global timestamp. This simple operation provides two crucial advantages: (i) CPA effectively mitigates inter‑variate asynchrony by forcing all variates to share a common timeline, thereby enabling the extraction of inter‑series correlations under a unified temporal resolution, and (ii) CPA transforms the variate‑length univariate series within an IMTS into fixed‑size representations, preserving observing sparsity while facilitating batch training and efficient processing by sequence modeling architectures, e.g., RNN \citep{Che2018} and Transformer \citep{Warpformer}.

Unfortunately, despite the aforementioned advantages, CPA however suffers from poor efficiency. This inefficiency arises because CPA must identify all global timestamps in an IMTS and indiscriminately pad every variate, which inevitably inflates the average sequence length. As shown in Fig.~\ref{fig:cpa}, each variate doubles from four to eight observations after alignment, leading to prohibitive computational overhead and memory bottlenecks, particularly when the number of variates is large \citep{tPatchGNN}. Recently, several graph‑based approaches have been proposed to bypass the length‑explosion issue introduced by CPA. For example, tPatchGNN \citep{tPatchGNN} chunks an IMTS into fixed‑length patches and pads each patch individually, but the rigid patch size distorts local temporal patterns. GraFITi \citep{GraFITi} and TimeCHEAT \citep{TimeCHEAT} treat the series as a bipartite graph; yet their message‑passing schemes cannot exchange information between variates that never co‑occur in time \citep{HyperIMTS}. 

Considering CPA’s unparalleled capacity in mitigating inter‑variate asynchrony and standardizing sequence length, \textbf{we are the first to argue that CPA should not be discarded as recent graph‑based detours have done, but should instead be fully exploited, provided that its efficiency issue can be properly resolved.} Therefore, we propose a new model named \textbf{KAFNet} for IMTS forecasting, proactively embracing CPA but mitigating its inefficiency. 

In KAFNet, a Pre‑Convolution layer first processes the pre‑aligned series for primary feature extraction by smoothing the length‑inflated series and enhancing local temporal patterns. Because the pre‑aligned series remains long, we then route it to the Temporal \textbf{K}ernel \textbf{A}ggregation (TKA) module, where a learnable bank of Gaussian kernels softly partitions the normalized timestamps and pools observations inside each partition, compressing the pre-aligned series while modeling the intra-series irregularity in a channel-independent way. The collection of per‑variate representations is fed to stacked \textbf{F}requency Linear Attention (FLA) blocks. Each FLA block integrates a frequency-enhanced linear attention mechanism with random fourier feature projection, effectively capturing inter-variate correlations with minimal computational overhead. Finally, the encoded representation is passed through a lightweight MLP-based output layer that enables query‑specific forecasts, after which the entire model is trained end‑to‑end with a standard mean‑squared‑error objective. To the best of our knowledge, we provide the first study directly addressing the inefficient series‑length explosion issue in CPA for IMTS forecasting. Our contributions can be articulated as follows:

\begin{figure}[t]
    \centering
    \includegraphics[width=0.9\linewidth, trim=0 30 0 23, clip]{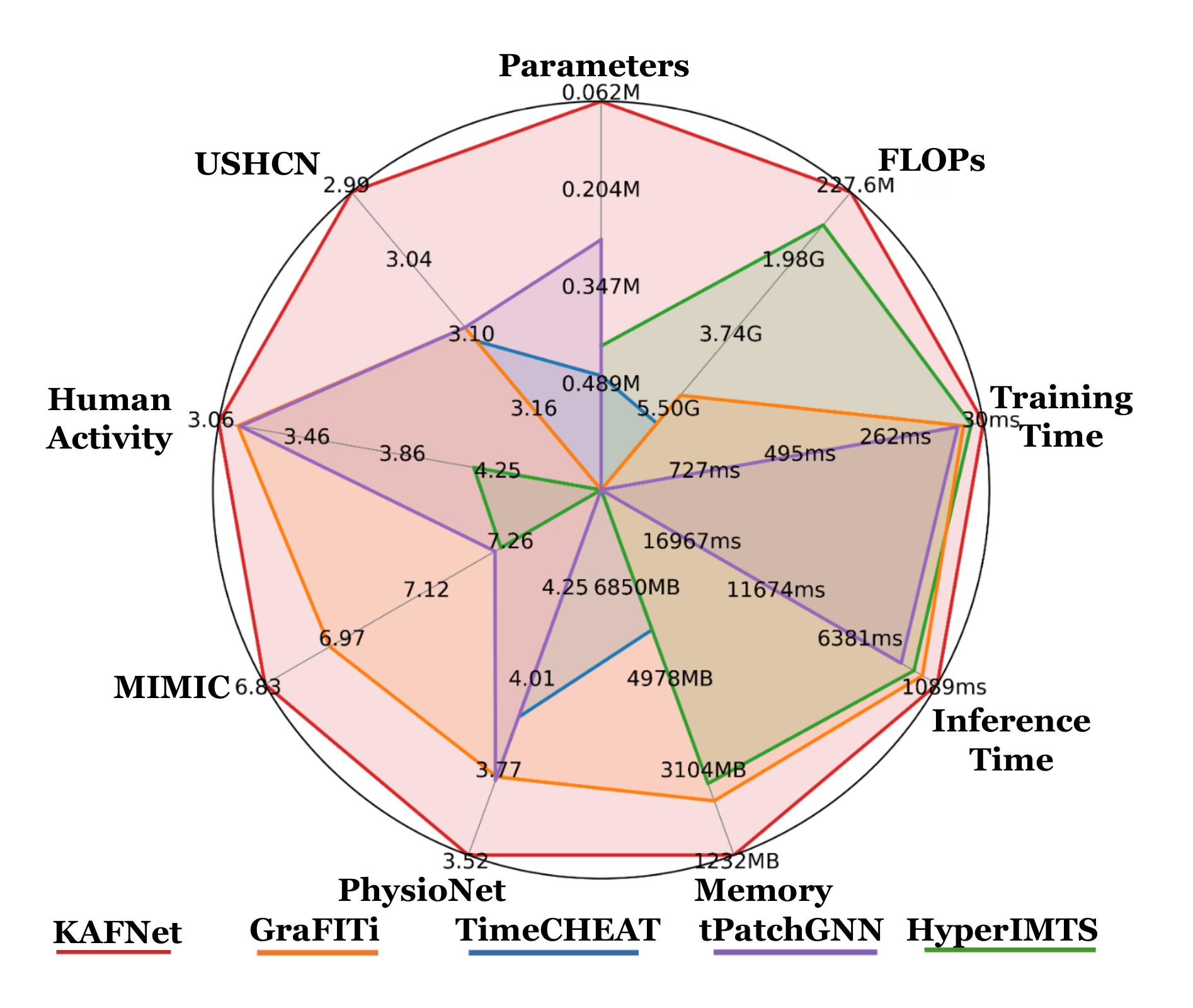}
    \caption{KAFNet delivers superior predictive accuracy (MAE) and efficiency (average) on four IMTS datasets.}
    \label{fig:radar}
\end{figure}

\begin{itemize}[leftmargin=*]
    \item  We revitalize Canonical Pre-Alignment for irregular multivariate time series forecasting by showing that, once the efficiency issue is addressed, model leverages CPA can outperform recent dominating graph-based baselines.
    \item  We introduce KAFNet for IMTS forecasting that seamlessly integrates a Pre‑Convolution module for sequence smoothing, a Temporal Kernel Aggregation module for temporal compression, and Frequency Linear Attention blocks for inter-variate correlations modeling, yielding an approach that is both compact and expressive.
    \item Extensive experiments on four public IMTS benchmark datasets show that KAFNet achieves superior predictive accuracy, realizing on average a \textbf{7.2$\times$} parameter reduction and a \textbf{8.4$\times$} training-inference acceleration compared with SOTA graph-based models (as shown in Fig.~\ref{fig:radar}).
\end{itemize}

\section{Related Works}

\subsection{IMTS Forecasting}

Deep learning for IMTS broadly follows two lines. (i) Continuous-time models parameterize dynamics with ODE/SDEs, including Neural ODE/Latent ODE \citep{Neural-ODE,Latent-ODE}, CRU \citep{CRU}, and GRU-ODE \citep{GRU-ODE}, and variants that reduce solver cost such as Neural Flows \citep{Neural-Flows}; ContiFormer \citep{ContiFormer} further couples Neural ODE dynamics with attention. These methods capture irregular trajectories but can be computationally heavy due to numerical solvers. (ii) Relational/patch-based methods avoid dense alignment by operating on patches or graphs, e.g., GraFITi \citep{GraFITi}, tPatchGNN \citep{tPatchGNN}, Hi-Patch \citep{Hi-Patch}, and HyperIMTS \citep{HyperIMTS}. While they alleviate length growth, their locality and message-passing often limit global inter-variate correlation modeling under a shared temporal grid. Our approach differs: we retain CPA for a unified timeline, but amortize its cost by learnable temporal compression and linear-time inter-variate modeling in the frequency domain.


\begin{figure*}[t] 
\centering \includegraphics[width=0.87\linewidth]{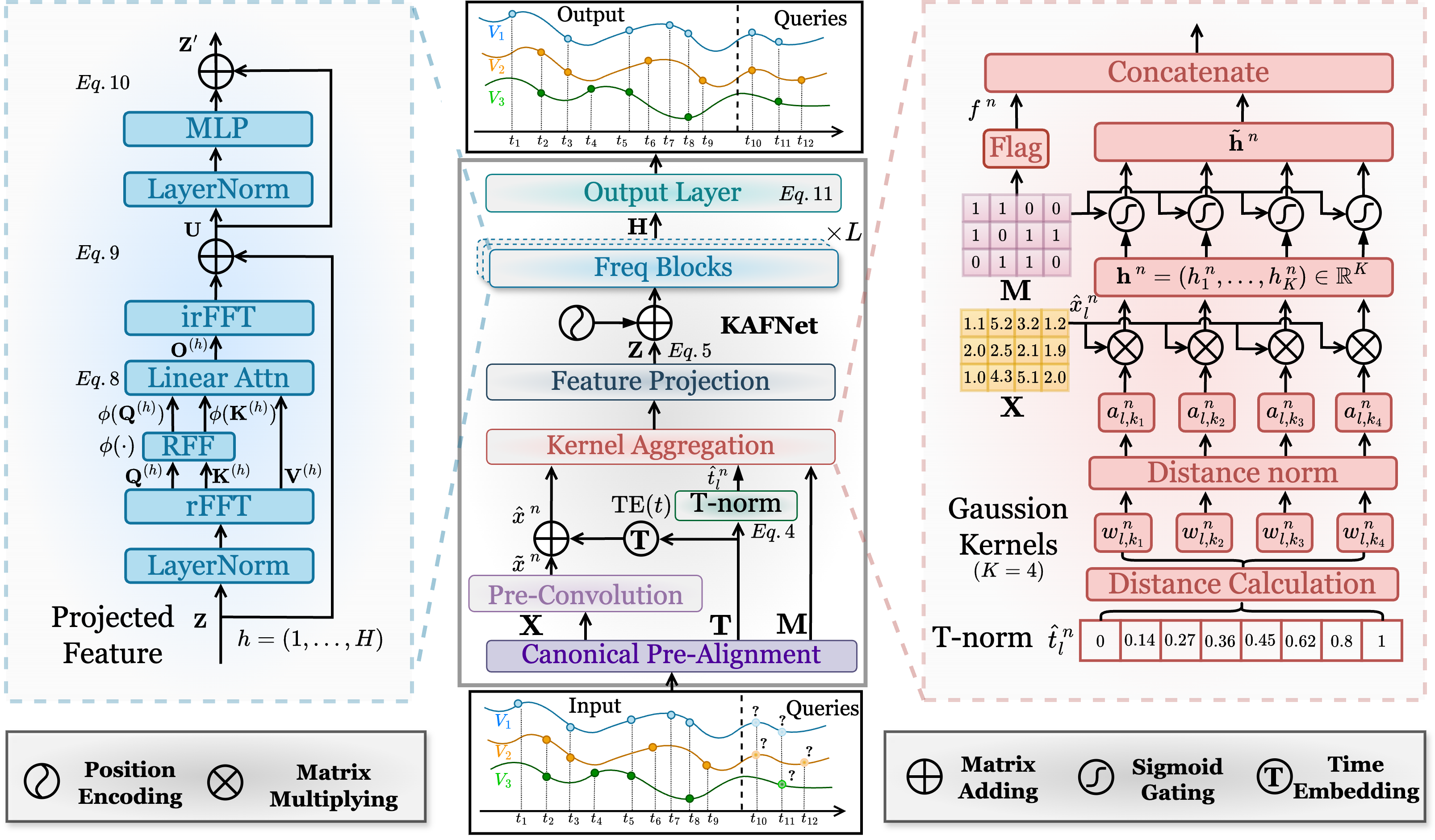} 
\caption{The main architecture of KAFNet. The input IMTS is initially processed by CPA and fed into the Pre‑Convolution module ($n\in[1,N]$) for sequence smoothing, then passed through the Temporal Kernel Aggregation module for intra‑series irregularity modeling and through the Frequency Linear Attention blocks for the inter‑series correlations modeling. Finally, the Output Layer generates the query-specific forecasts. Linear Attn: linear attention mechanism, MLP: multi-layer perceptron.} \label{fig:arch} 
\end{figure*}

\subsection{CPA for IMTS Analysis}
CPA aligns all variates onto a shared grid, filling missing values and recording availability with a mask; it is widely adopted across IMTS tasks, including GRU-D \citep{Che2018}, Latent ODE \citep{Latent-ODE}, mTAND \citep{mTAND}, Neural Flows \citep{Neural-Flows}, CSDI \citep{CSDI}, RainDrop \citep{RainDrop}, and Warpformer \citep{Warpformer}. The benefit is fixed-shape inputs and mitigated inter-variate asynchrony; the drawback is inflated sequence length $L$, which raises memory and compute, especially with many variates \citep{tPatchGNN}. Recent work mitigates this via patch-based alignment (e.g., tPatchGNN \citep{tPatchGNN}, APN \citep{APN}) built upon PatchTST-style slicing \citep{PatchTST}, but per-variate patching can distort local patterns and overlook cross-variate correlations. In contrast, we keep full CPA and address its inefficiency directly through Temporal Kernel Aggregation (for compression) and Frequency Linear Attention (for global inter-variate modeling).

\section{Preliminary}
\label{sec:preliminary}

\subsection{Problem Formulation}
\label{subsec:problem_formulation}
\subsubsection{Definition (Irregular Multivariate Time Series).}
An IMTS with $N$ variates is commonly expressed as \(\mathcal{O}=\{[(t_i^{n},x_i^{n})]_{i=1}^{L_n}\}_{n=1}^{N}\). For the $n$-th variate, \((t_i^{n},x_i^{n})\in\mathbb{R}^2\) represents the value \(x_i^{n}\) recorded at time \(t_i^{n}\). Observation counts \(L_n\) can vary between variates. The intervals are typically non-uniform, indicating intra-variate irregularity. Additionally, timestamps across different variates do not align, creating inter-variate asynchrony.

\subsubsection{Problem (Irregular Multivariate Time Series Forecasting).} Let \(\mathcal{Q}=\{[q_j^{n}]_{j=1}^{Q_n}\}_{n=1}^{N}\) denote the collection of future query timestamps, where \(q_j^{n}>\max_{i}t_i^{n}\) and \(Q_n\) is the number of queries on variate \(n\).  The forecasting task targets on learning a model \(\mathcal{F}_\theta(\cdot,\cdot)\), parameterized by \(\theta\), that maps historical observations \(\mathcal{O}\) and queries \(Q\) to predictions, \(\mathcal{F}_\theta(\mathcal{O},\mathcal{Q})\rightarrow\hat{\mathcal{X}}=\{[\hat{x}_j^{n}]_{j=1}^{Q_n}\}_{n=1}^{N}\), where \(\hat{x}_j^{n}\) approximates the predicted future value at time \(q_j^{n}\).

\subsection{Canonical Pre-Alignment}
\label{subsec:prealignment}
Canonical Pre-Alignment is a standard data preprocessing technique to alleviate inter-variate asynchrony \cite{Che2018,Latent-ODE,mTAND,Neural-Flows,CSDI,RainDrop,Warpformer}. Specifically, IMTS data \(\mathcal{O}\) is transformed into a triplet \((\mathcal{T},\mathcal{X},\mathcal{M})\). First, \(\mathcal{T}=[t_l]_{l=1}^{L}\in\mathbb{R}^{L}\) is obtained by merging all timestamps and sorting them, i.e., \ \(\mathcal{T}= \bigcup_{n=1}^{N}\{t_i^{n}\}_{i=1}^{L_n}\). Second, the value matrix \(\mathcal{X}=[[x_l^{n}]_{n=1}^{N}]_{l=1}^{L}\in\mathbb{R}^{L\times N}\) assigns \(x_l^{n}=x_i^{n}\) if a matching observation exists, otherwise a placeholder such as zero is stored. Finally, the binary mask \(\mathcal{M}=[[m_l^{n}]_{n=1}^{N}]_{l=1}^{L}\in\{0,1\}^{L\times N}\) with \(m_l^{n}=1\) when \(x_l^{n}\) is observed and \(0\) otherwise. This aligned grid preserves the irregular sampling pattern through \(\mathcal{M}\) while enabling subsequent models to process fixed-shape inputs. However, this technique may significantly increase the average sequence length, leading to scalability concerns particularly when modeling IMTS with a large number of variates \cite{tPatchGNN}.

\section{Methodology}
\label{sec:methodology}

Our model is composed of four modular components (Fig. \ref{fig:arch}). (i) \textbf{Pre‑Convolution} module, which rectifies the uneven information distribution introduced by CPA’s zero‑padding. (ii) \textbf{Temporal Kernel Aggregation} module, which captures temporal irregularities and compresses each lengthy, pre‑aligned sequence into a concise representation. (iii) a series of \textbf{Frequency Linear Attention} blocks (Freq Blocks), which efficiently models inter‑variate dependencies in the frequency domain. (iv) MLP‑based \textbf{Output Layer}, which addresses arbitrary forecasting queries. The entire architecture is trained end‑to‑end by minimizing the mean squared error. In the following, we use \(\mathbf{T},\mathbf{X},\mathbf{M}\in\mathbb{R}^{L\times N}\) to represent the triplet \((\mathcal{T},\mathcal{X},\mathcal{M})\). From \((\mathbf{T},\mathbf{X},\mathbf{M})\),
we have
a length-\(L\) time series \(x^{n}=(x_{1}^{n},\dots ,x_{L}^{n})\in\mathbb{R}^{L}\)
for each variate $n\in[1,N]$.
In the sequel, we will use \(\!l\!\in\![1,L]\!\) to index the time grid (observation).

\subsection{Pre-Convolution for Sequence Smoothing}
In CPA, zeros are introduced for padding to unify all time steps to a common time axis. However, zero-padding triggers a severe imbalance in information distribution within the pre-aligned series. Therefore, before feeding the input pre-aligned series \(x^{n}\) into complex architecture for deep representation learning, we propose preprocessing the sequence to reduce information sparsity while enhancing its smoothness. Specifically, we pass \(x^{n}\) through two lightweight convolutions acting along the time dimension:
\begin{equation}
  \tilde{x}^{n}
  =\operatorname{Conv}_{1\times1}\!
     \bigl(\sigma\bigl(\operatorname{Conv}_{1\times3}(x^{n})\bigr)\bigr)\in\mathbb{R}^{L}, \label{eq:preconv}
\end{equation}
where \(\sigma\) is \texttt{ReLU}. For the same variate \(n\), let
\(
  \mathbf{t}^{n}=(t_{1}^{n},\dots ,t_{L}^{n})
\)
denote the CPA-generated timestamps in \(\mathbf{T}\).
Given that the time grid indices generated by CPA no longer reflect true elapsed time, we encode continuous temporal information with a time embedding function defined for a scalar timestamp \(t\):
\begin{equation}
  \mathrm{TE}(t)=\bigl[w_{s}t+b_{s}\oplus\sin(\mathbf{w}_{p}t+\mathbf{b}_{p})\oplus\cos(\mathbf{w}_{c}t+\mathbf{b}_{c})\bigr], \label{eq:te}
\end{equation}
which is applied element-wise to \(\mathbf{t}^{n}\), i.e., to each \(t_{l}^{n}\).
Finally, after a linear projection with a learnable vector \(\mathbf{w}_{t}\in\mathbb{R}^{d_{\mathrm{te}}}\), we obtain the time-aware encoded representation:
\begin{equation}
  \hat{x}^{n}
  =\tilde{x}^{n}+\mathbf{w}_{t}^{\top}\mathrm{TE}(\mathbf{t}^{n})
  \;\in\mathbb{R}^{L}, \label{eq:te_fusion}
\end{equation}
where \(\mathbf{w}_{t}^{\top}\mathrm{TE}(\mathbf{t}^{n})\) denotes the length-\(L\) vector whose \(l\)-th element is \(\mathbf{w}_{t}^{\top}\mathrm{TE}(t_{l}^{n})\).
This representation is then sent to the subsequent TKA module for explicit temporal-irregularity modeling and sequence compression.

\subsection{Temporal Kernel Aggregation (TKA)}
As analyzed in \citep{tPatchGNN}, CPA inevitably increases the average length of an IMTS, thereby making the modeling of the extended series less efficient. Therefore, we propose a new Temporal Kernel Aggregation (TKA) module to significantly reduce the series length while explicitly encoding the intra-series irregularity. 

Specifically, after Pre-Convolution, each variate is represented by \(\hat{x}_{l}^{n}\in\mathbb{R}^{L}\). Here, the subscript \(l\) denotes the time-grid index (observation). To obtain a fixed-length embedding of the input pre-aligned series while modeling the temporal irregularity, we first map every time index in \(\hat{x}_{l}^{n}\) to the unit interval by min-max normalization:
\begin{equation}
\hat{t}_{l}^{n}=\frac{t_{l}^{n}-t_{\min}^{n}}{t_{\max}^{n}-t_{\min}^{n}}\in[0,1],
\end{equation}
where $t_{\min}^{n}$ and $t_{\max}^{n}$ denote the first and last canonical timestamps for variate $n$, respectively. Subsequently, on this normalized axis, we place \(K\) Gaussian kernels whose centers \(c_{k}\)'s are evenly spaced and whose bandwidths \(\{\sigma_{k}=e^{\log\alpha_{k}}\}\) are learnable, so that together they define a smooth partition of the timeline. Intuitively, these \(K\) Gaussian kernels form a soft temporal codebook on the normalized time axis, with each timestamp softly assigned to nearby codewords according to its Gaussian affinity. The closeness of the \(l\)-th timestamp 
$\hat{t}_{l}^{n}$
to kernel \(k\) is measured by
\(
w_{l,k}^{n}=\exp\bigl[-\tfrac12(\hat{t}_{l}^{n}-c_{k})^{2}/\sigma_{k}^{2}\bigr]\,m_{l}^{n},
\)
where $m_{l}^{n}$ is the binary mask indicating whether $x_{l}^{n}$ is observed. Normalization then produces the coefficient:
\(
a_{l,k}^{n}=\frac{w_{l,k}^{n}}{\sum_{j=1}^{L}w_{j,k}^{n}}, 
\)
which quantifies how strongly the \(l\)-th observation contributes to the \(k\)-th temporal region. Then, we stack \(a_{l,k}^{n}\) over \(l\) and \(k\), pooling the time series values \(\{\hat{x}_{l}^{n}\}\) according to these coefficients through
\(
h_{k}^{n}=\sum_{l=1}^{L}a_{l,k}^{n}\,\hat{x}_{l}^{n}\), forming \(\mathbf{h}^{n}=(h_{1}^{n},\dots,h_{K}^{n})\in\mathbb{R}^{K}
\). This operation gives one summary per kernel, so \(\mathbf{h}^{n}\) captures the signal present in \(K\) smooth temporal windows. A learnable gate vector \(\mathbf{g}\in\mathbb{R}^{K}\) modulates their importance by the element-wise product \(\tilde{\mathbf{h}}^{n}=Sigmoid(\mathbf{g})\odot\mathbf{h}^{n}\). Then, the binary flag \(f^{n}=\mathbb{I}(\sum_{l}m_{l}^{n}>0)\) is concatenated as \(\tilde{\mathbf{h}}^{n}\oplus f^{n}\in\mathbb{R}^{K+1}\). Finally, the Feature Projection module in Fig. \ref{fig:arch} using a linear layer weighted 
\(
\mathbf{W}_{\mathrm{proj}}\in\mathbb{R}^{(K+1)\times d}
\)
with the hidden state dimension of \(d\) projects the vector through:
\begin{equation}
\mathbf{z}^{n}=(\tilde{\mathbf{h}}^{n}\oplus f^{n})\mathbf{W}_{\mathrm{proj}}\in\mathbb{R}^{d}, \label{eq:tka_proj}
\end{equation}
yielding a compact embedding whose size is independent of \(L\) yet still encodes the original irregular timing through the Gaussian weighting mechanism. Stacking all \(\mathbf{z}^{n}\)'s forms \(\mathbf{Z}=[\mathbf{z}^{1},\dots,\mathbf{z}^{N}]\in\mathbb{R}^{N\times d}\), which serves as the input to the subsequent frequency-domain blocks. 

\subsection{Frequency Linear Attention (FLA) Blocks}
\label{subsec:freq_block_new}
Transforming the representation into the frequency domain has been proved effective in capturing inter-variate
dependencies and global information \citep{AirRadar, CirT}. Therefore, we propose multi‑layer Frequency Linear Attention (FLA) blocks that capture inter-variate correlations by employing a linearized attention mechanism in the frequency domain, maintaining low computational complexity. Let \(\mathbf{Z}\in\mathbb{R}^{N\times d}\) denote the input representation where each row corresponds to one variate obtained from the TKA. In each block layer, a LayerNorm first normalizes the input, and a real-valued FFT (rFFT) is applied along the hidden dimension of \(\mathbf{Z}\) to convert \(\mathbf{Z}\) into frequency coefficients:
\begin{equation}
  \mathbf{C} = \operatorname{\textbf{rFFT}}\bigl(\mathrm{LayerNorm}(\mathbf{Z})\bigr) \in \mathbb{R}^{N \times 2d_f}, d_f = d/2, 
  \label{eq:fft}
\end{equation}
where \(d\) is the hidden state dimension. Subsequently, FLA leverages multi-head self-attention mechanism to capture inter-variate correlation. Each head \(h=(1,\dots,H)\) computes its query, key, and value matrices by applying learned projections to the shared frequency representation through:
\begin{equation}
  \mathbf{Q}^{(h)}=\mathbf{C}\mathbf{W}^{Q}_{h},
  \mathbf{K}^{(h)}=\mathbf{C}\mathbf{W}^{K}_{h},
  \mathbf{V}^{(h)}=\mathbf{C}\mathbf{W}^{V}_{h},
  \label{eq:qkv}
\end{equation}
where \(\mathbf{W}^{Q}_{h}, \mathbf{W}^{K}_{h}, \mathbf{W}^{V}_{h}\in \mathbb{R}^{2d_f \times d_h}\), and \(d_h = 2d_f / H\).

To avoid the quadratic cost of computing \(\exp(\mathbf{Q}^{\top}\mathbf{K})\) in Softmax attention \citep{comba}, we approximate the Softmax kernel using a Random Fourier Feature (RFF) \citep{rff} map \(\phi(\cdot)\), defined as:
\(
  \!\phi(\mathbf{x})\! \!= \!\!\tfrac{1}{\sqrt{R}} \bigl[\cos(\mathbf{\Omega}^{\!\top}\mathbf{x} + \mathbf{b}),\sin(\mathbf{\Omega}^{\!\top}\mathbf{x} + \mathbf{b})\bigr] \!\!\in\! \!\mathbb{R}^{R}\!,
\)
where \(\mathbf{\Omega} \!\in\! \mathbb{R}^{d_h \times R/2}\!\), \(\!\mathbf{b} \!\in \!\mathbb{R}^{R/2}\!\) are randomly initialized, yielding a closed-form attention weight computation with linear complexity:
\begin{equation}
  \mathbf{O}^{(h)} = 
  \frac{\phi(\mathbf{Q}^{(h)}) \left(\phi(\mathbf{K}^{(h)})^\top \mathbf{V}^{(h)}\right)}
       {\phi(\mathbf{Q}^{(h)}) \left(\phi(\mathbf{K}^{(h)})^\top \right)}.
  \label{eq:linear_attn}
\end{equation}

Subsequently, all multi-head outputs \(\mathbf{O}^{(1)},\dots,\mathbf{O}^{(H)}\) are concatenated and transformed back to its original domain via inverse real FFT (irFFT). A residual connection with the input is then applied:
\begin{equation}
  \mathbf{U} = \mathbf{Z} + \operatorname{\textbf{irFFT}}\bigl([\mathbf{O}^{(1)}\;\!\!\dots\;\!\!\mathbf{O}^{(H)}]\bigr).
  \label{eq:ifft_res}
\end{equation}
To further enhance representational capacity, \(\mathbf{U}\) is normalized again and passed through a feed-forward network (two-layer MLP), followed by a second residual connection:
\begin{equation}
  \mathbf{Z}' = \mathbf{U} + \mathrm{MLP}(\mathrm{LayerNorm}(\mathbf{U})). \label{eq:mlp_res}
\end{equation}

The above operations define one Frequency Linear Attention block. By stacking \(L\) such blocks, we obtain the final inter-variate representation \(\mathbf{Z}^{(L)}\), which is further projected by a learnable matrix \(\mathbf{W}_{a} \in \mathbb{R}^{d \times d}\) to produce the output of the entire frequency attention module:
\(
  \mathbf{H} = \mathbf{W}_{a} \mathbf{Z}^{(L)}.
  \label{eq:freq_output}
\)
The proposed FLA blocks enable efficient and expressive modeling of dependencies among variates.

\subsection{Output Layer and Training Objective}
\label{subsec:decoder}
After temporal irregularity has been compressed by TKA and inter-variate correlations have been refined by the \(L\) stacked FLA blocks, KAFNet outputs the encoded representation \(\mathbf{H}\in\mathbb{R}^{N\times d}\); its \(n\)-th row \(\mathbf{H}^{n}\) is a compact summary of variate \(n\). For each query time \(q_{j}^{n}\), we concatenate this summary with its time embedding and directly map the result to the scalar prediction via a three-layer MLP:
\begin{equation}
  \hat{x}_{j}^{n}
  =\text{MLP}\Bigl(\,\mathbf{H}^{n}\oplus\mathrm{TE}(q_{j}^{n})\Bigr).
  \label{eq:forecast}
\end{equation}

On training, we minimize the mean-squared error (MSE) over all variates and their query points:
\begin{equation}
\mathcal{L}= 
\frac{1}{N}\sum_{n=1}^{N}
\frac{1}{Q_{n}}\sum_{j=1}^{Q_{n}}\bigl(\hat{x}_{j}^{n}-x_{j}^{n}\bigr)^{2}.
\label{eq:loss}
\end{equation}
Optimizing the parameters in Eq.~\eqref{eq:preconv} to Eq.~\eqref{eq:forecast} thus jointly learns temporal abstraction, inter-variate correlation, and query-aware prediction in an end-to-end manner.

\subsection{Computational Complexity}
\label{subsec:complexity}
Per forward pass, computation proceeds through four stages on a pre-aligned length-\(L\) series with \(N\) variates: (i) Pre-Conv, (ii) TKA that compresses each variate to \(d\) features, (iii) a length-invariant FLA block, and (iv) the output head. The total cost is
\[
\begin{aligned}
\Omega
&= \underbrace{4NLd}_{\smash{\text{Pre-Conv}}}
 + \underbrace{N(3LK{+}Kd)}_{\smash{\text{TKA}}}
\\
&\quad + \underbrace{2Nd\log d + 3Nd^{2} + 2NdR}_{\smash{\text{FLA}}}
 + \underbrace{QNd^{2}}_{\smash{\text{Output}}}
\\
&= N\!\left[(4d{+}3K)L + Kd + (Q{+}3)d^{2} + 2d(\log d + R)\right].
\end{aligned}
\]
Only \((4d{+}3K)L\) depends on \(L\). Thus, \(\Omega\) scales linearly in both \(L\) and \(N\), while post-TKA computation (FLA and output head) is length-invariant due to TKA’s compression.

\section{Experiments}

\begin{table*}[t]
  \centering
  \scalebox{0.77}{
  \setlength{\tabcolsep}{3pt}
  \begin{tabular}{c | cc | cc | cc | cc | c}
    \toprule
    \textbf{Dataset} & \multicolumn{2}{c|}{\textbf{PhysioNet}} & \multicolumn{2}{c|}{\textbf{MIMIC}} & \multicolumn{2}{c|}{\textbf{Human Activity}} & \multicolumn{2}{c|}{\textbf{USHCN}} & \textbf{Average}\\
    \cmidrule(lr){1-9}
    \textbf{Metric}& MSE$\times 10^{-3}$ & MAE$\times 10^{-2}$ & MSE$\times 10^{-2}$ & MAE$\times 10^{-2}$ & MSE$\times 10^{-3}$ & MAE$\times 10^{-2}$ & MSE$\times 10^{-1}$ & MAE$\times 10^{-1}$ & \textbf{Rank}\\
    \midrule
    DLinear (2023)& 41.86 {\scriptsize $\pm$ 0.05}$_{(23)}$ & 15.52 {\scriptsize $\pm$ 0.03}$_{(23)}$ & 4.90 {\scriptsize $\pm$ 0.00}$_{(23)}$ & 16.29 {\scriptsize $\pm$ 0.05}$_{(23)}$ & 4.03 {\scriptsize $\pm$ 0.01}$_{(14)}$ & 4.21 {\scriptsize $\pm$ 0.01}$_{(15)}$ & 6.21 {\scriptsize $\pm$ 0.00}$_{(23)}$ & 3.88 {\scriptsize $\pm$ 0.02}$_{(23)}$ & 20.9 \\
    TimesNet (2023)& 16.48 {\scriptsize $\pm$ 0.11}$_{(22)}$ & 6.14 {\scriptsize $\pm$ 0.03}$_{(22)}$ & 5.88 {\scriptsize $\pm$ 0.08}$_{(22)}$ & 13.62 {\scriptsize $\pm$ 0.07}$_{(22)}$ & 3.12 {\scriptsize $\pm$ 0.01}$_{(11)}$ & 3.56 {\scriptsize $\pm$ 0.02}$_{(11)}$ & 5.58 {\scriptsize $\pm$ 0.05}$_{(14)}$ & 3.60 {\scriptsize $\pm$ 0.04}$_{(18)}$ & 17.8 \\
    PatchTST (2023)& 12.00 {\scriptsize $\pm$ 0.23}$_{(21)}$ & 6.02 {\scriptsize $\pm$ 0.14}$_{(21)}$ & 3.78 {\scriptsize $\pm$ 0.03}$_{(21)}$ & 12.43 {\scriptsize $\pm$ 0.10}$_{(21)}$ & 4.29 {\scriptsize $\pm$ 0.14}$_{(17)}$ & 4.80 {\scriptsize $\pm$ 0.09}$_{(18)}$ & 5.75 {\scriptsize $\pm$ 0.01}$_{(17)}$ & 3.57 {\scriptsize $\pm$ 0.02}$_{(17)}$ & 19.1\\
    Crossformer (2023)& 6.66 {\scriptsize $\pm$ 0.11}$_{(13)}$ & 4.81 {\scriptsize $\pm$ 0.11}$_{(16)}$ & 2.65 {\scriptsize $\pm$ 0.10}$_{(17)}$ & 9.56 {\scriptsize $\pm$ 0.29}$_{(17)}$ & 4.29 {\scriptsize $\pm$ 0.20}$_{(18)}$ & 4.89 {\scriptsize $\pm$ 0.17}$_{(19)}$ & 5.25 {\scriptsize $\pm$ 0.04}$_{(7)}$ & 3.27 {\scriptsize $\pm$ 0.09}$_{(11)}$ & 14.8\\
    Graph Wavenet (2019)& 6.04 {\scriptsize $\pm$ 0.28}$_{(8)}$ & 4.41 {\scriptsize $\pm$ 0.11}$_{(9)}$ & 2.93 {\scriptsize $\pm$ 0.09}$_{(19)}$ & 10.50 {\scriptsize $\pm$ 0.15}$_{(19)}$ & 2.89 {\scriptsize $\pm$ 0.03}$_{(6)}$ & 3.40 {\scriptsize $\pm$ 0.05}$_{(6)}$ & 5.29 {\scriptsize $\pm$ 0.04}$_{(9)}$ & 3.16 {\scriptsize $\pm$ 0.09}$_{(6)}$ & 10.3\\
    MTGNN (2020)& 6.26 {\scriptsize $\pm$ 0.18}$_{(11)}$ & 4.46 {\scriptsize $\pm$ 0.07}$_{(10)}$ & 2.71 {\scriptsize $\pm$ 0.23}$_{(18)}$ & 9.55 {\scriptsize $\pm$ 0.65}$_{(16)}$ & 3.03 {\scriptsize $\pm$ 0.03}$_{(9)}$ & 3.53 {\scriptsize $\pm$ 0.03}$_{(9)}$ & 5.39 {\scriptsize $\pm$ 0.05}$_{(12)}$ & 3.34 {\scriptsize $\pm$ 0.02}$_{(12)}$ & 12.1\\
    StemGNN (2020)& 6.86 {\scriptsize $\pm$ 0.28}$_{(15)}$ & 4.76 {\scriptsize $\pm$ 0.19}$_{(15)}$ & 1.73 {\scriptsize $\pm$ 0.02}$_{(7)}$ & 7.71 {\scriptsize $\pm$ 0.11}$_{(9)}$ & 8.81 {\scriptsize $\pm$ 0.37}$_{(21)}$ & 6.90 {\scriptsize $\pm$ 0.02}$_{(21)}$ & 5.75 {\scriptsize $\pm$ 0.09}$_{(18)}$ & 3.40 {\scriptsize $\pm$ 0.09}$_{(13)}$ & 14.9\\
    CrossGNN (2023)& 7.22 {\scriptsize $\pm$ 0.36}$_{(17)}$ & 4.96 {\scriptsize $\pm$ 0.12}$_{(17)}$ & 2.95 {\scriptsize $\pm$ 0.16}$_{(20)}$ & 10.82 {\scriptsize $\pm$ 0.21}$_{(20)}$ & 3.03 {\scriptsize $\pm$ 0.10}$_{(10)}$ & 3.48 {\scriptsize $\pm$ 0.08}$_{(8)}$ & 5.66 {\scriptsize $\pm$ 0.04}$_{(16)}$ & 3.53 {\scriptsize $\pm$ 0.05}$_{(15)}$ & 15.4\\
    FourierGNN (2023)& 6.84 {\scriptsize $\pm$ 0.35}$_{(14)}$ & 4.65 {\scriptsize $\pm$ 0.12}$_{(13)}$ & 2.55 {\scriptsize $\pm$ 0.03}$_{(16)}$ & 10.22 {\scriptsize $\pm$ 0.08}$_{(18)}$ & 2.99 {\scriptsize $\pm$ 0.02}$_{(8)}$ & 3.42 {\scriptsize $\pm$ 0.02}$_{(7)}$ & 5.82 {\scriptsize $\pm$ 0.06}$_{(21)}$ & 3.62 {\scriptsize $\pm$ 0.07}$_{(20)}$ & 14.6\\
    \midrule
    GRU-D (2018)& 5.59 {\scriptsize $\pm$ 0.09}$_{(4)}$ & 4.08 {\scriptsize $\pm$ 0.05}$_{(6)}$ & 1.76 {\scriptsize $\pm$ 0.03}$_{(9)}$ & 7.53 {\scriptsize $\pm$ 0.09}$_{(7)}$ & 2.94 {\scriptsize $\pm$ 0.05}$_{(7)}$ & 3.53 {\scriptsize $\pm$ 0.06}$_{(10)}$ & 5.54 {\scriptsize $\pm$ 0.38}$_{(13)}$ & 3.40 {\scriptsize $\pm$ 0.28}$_{(14)}$ & 8.8\\
    SeFT (2020)& 9.22 {\scriptsize $\pm$ 0.18}$_{(19)}$ & 5.40 {\scriptsize $\pm$ 0.08}$_{(19)}$ & 1.87 {\scriptsize $\pm$ 0.01}$_{(11)}$ & 7.84 {\scriptsize $\pm$ 0.08}$_{(11)}$ & 12.20 {\scriptsize $\pm$ 0.17}$_{(22)}$ & 8.43 {\scriptsize $\pm$ 0.07}$_{(22)}$ & 5.80 {\scriptsize $\pm$ 0.19}$_{(20)}$ & 3.70 {\scriptsize $\pm$ 0.11}$_{(22)}$ & 18.3\\
    RainDrop (2021)& 9.82 {\scriptsize $\pm$ 0.08}$_{(20)}$ & 5.57 {\scriptsize $\pm$ 0.06}$_{(20)}$ & 1.99 {\scriptsize $\pm$ 0.03}$_{(15)}$ & 8.27 {\scriptsize $\pm$ 0.07}$_{(15)}$ & 14.92 {\scriptsize $\pm$ 0.14}$_{(23)}$ & 9.45 {\scriptsize $\pm$ 0.05}$_{(23)}$ & 5.78 {\scriptsize $\pm$ 0.22}$_{(19)}$ & 3.67 {\scriptsize $\pm$ 0.17}$_{(21)}$ & 19.5\\
    Warpformer (2023)& 5.94 {\scriptsize $\pm$ 0.35}$_{(6)}$ & 4.21 {\scriptsize $\pm$ 0.12}$_{(7)}$ & 1.73 {\scriptsize $\pm$ 0.04}$_{(8)}$ & 7.58 {\scriptsize $\pm$ 0.13}$_{(8)}$ & 2.79 {\scriptsize $\pm$ 0.04}$_{(5)}$ & 3.39 {\scriptsize $\pm$ 0.03}$_{(5)}$ & 5.25 {\scriptsize $\pm$ 0.05}$_{(8)}$ & 3.23 {\scriptsize $\pm$ 0.05}$_{(8)}$ & 6.9\\
    \midrule
    mTAND (2021)& 6.23 {\scriptsize $\pm$ 0.24}$_{(10)}$ & 4.51 {\scriptsize $\pm$ 0.17}$_{(12)}$ & 1.85 {\scriptsize $\pm$ 0.06}$_{(10)}$ & 7.73 {\scriptsize $\pm$ 0.13}$_{(10)}$ & 3.22 {\scriptsize $\pm$ 0.07}$_{(12)}$ & 3.81 {\scriptsize $\pm$ 0.07}$_{(12)}$ & 5.33 {\scriptsize $\pm$ 0.05}$_{(10)}$ & 3.26 {\scriptsize $\pm$ 0.10}$_{(10)}$ & 10.8\\
    \midrule
    Latent-ODE (2019)& 6.05 {\scriptsize $\pm$ 0.57}$_{(9)}$ & 4.23 {\scriptsize $\pm$ 0.26}$_{(8)}$ & 1.89 {\scriptsize $\pm$ 0.19}$_{(13)}$ & 8.11 {\scriptsize $\pm$ 0.52}$_{(14)}$ & 3.34 {\scriptsize $\pm$ 0.11}$_{(13)}$ & 3.94 {\scriptsize $\pm$ 0.12}$_{(13)}$ & 5.62 {\scriptsize $\pm$ 0.03}$_{(15)}$ & 3.60 {\scriptsize $\pm$ 0.12}$_{(19)}$ & 13.0\\
    CRU (2022)& 8.56 {\scriptsize $\pm$ 0.26}$_{(18)}$ & 5.16 {\scriptsize $\pm$ 0.09}$_{(18)}$ & 1.97 {\scriptsize $\pm$ 0.02}$_{(14)}$ & 7.93 {\scriptsize $\pm$ 0.19}$_{(12)}$ & 6.97 {\scriptsize $\pm$ 0.78}$_{(20)}$ & 6.30 {\scriptsize $\pm$ 0.47}$_{(20)}$ & 6.09 {\scriptsize $\pm$ 0.17}$_{(22)}$ & 3.54 {\scriptsize $\pm$ 0.18}$_{(16)}$ & 17.5\\
    Neural Flow (2021)& 7.20 {\scriptsize $\pm$ 0.07}$_{(16)}$ & 4.67 {\scriptsize $\pm$ 0.04}$_{(14)}$ & 1.87 {\scriptsize $\pm$ 0.05}$_{(12)}$ & 8.03 {\scriptsize $\pm$ 0.19}$_{(13)}$ & 4.05 {\scriptsize $\pm$ 0.13}$_{(15)}$ & 4.46 {\scriptsize $\pm$ 0.09}$_{(16)}$ & 5.35 {\scriptsize $\pm$ 0.05}$_{(11)}$ & 3.25 {\scriptsize $\pm$ 0.05}$_{(9)}$ & 13.3\\
    tPatchGNN (2024)& \underline{4.98 {\scriptsize $\pm$ 0.08}}$_{(2)}$ & \underline{3.72 {\scriptsize $\pm$ 0.03}}$_{(2)}$ &  \underline{1.69 {\scriptsize $\pm$ 0.03}}$_{(2)}$ & 7.22 {\scriptsize $\pm$ 0.09}$_{(3)}$ & \underline{2.66 {\scriptsize $\pm$ 0.03}}$_{(2)}$ & 3.15 {\scriptsize $\pm$ 0.02}$_{(3)}$ & 5.00 {\scriptsize $\pm$ 0.04}$_{(3)}$ & \underline{3.08 {\scriptsize $\pm$ 0.04}}$_{(2)}$ & \underline{2.4}\\
    tPatchGNN* (2024)& 6.41 {\scriptsize $\pm$ 0.07}$_{(12)}$ & 3.89 {\scriptsize $\pm$ 0.07}$_{(5)}$ & 1.71 {\scriptsize $\pm$ 0.03}$_{(5)}$ & 7.43 {\scriptsize $\pm$ 0.11}$_{(6)}$ & 2.76 {\scriptsize $\pm$ 0.03}$_{(4)}$ & 3.23 {\scriptsize $\pm$ 0.04}$_{(4)}$ & 5.00 {\scriptsize $\pm$ 0.05}$_{(4)}$ & 3.09 {\scriptsize $\pm$ 0.03}$_{(3)}$ & 5.4\\
    GraFITi* (2024) & 6.02 {\scriptsize $\pm$ 0.06}$_{(7)}$ & 3.73 {\scriptsize $\pm$ 0.03}$_{(3)}$ & 1.71 {\scriptsize $\pm$ 0.02}$_{(4)}$ & \underline{6.94 {\scriptsize $\pm$ 0.03}}$_{(2)}$ & 2.73 {\scriptsize $\pm$ 0.03}$_{(3)}$ & \underline{3.14 {\scriptsize $\pm$ 0.02}}$_{(2)}$ & 5.09 {\scriptsize $\pm$ 0.03}$_{(5)}$ & 3.09 {\scriptsize $\pm$ 0.04}$_{(4)}$ & 3.8\\
    
    TimeCHEAT* (2025)& 5.05 {\scriptsize $\pm$ 0.08}$_{(3)}$ & 3.89 {\scriptsize $\pm$ 0.04}$_{(4)}$ & 1.70 {\scriptsize $\pm$ 0.01}$_{(3)}$ & 7.40 {\scriptsize $\pm$ 0.09}$_{(5)}$ & 4.06 {\scriptsize $\pm$ 0.08}$_{(16)}$ & 4.65 {\scriptsize $\pm$ 0.04}$_{(17)}$ & \textbf{4.42 {\scriptsize $\pm$ 0.04}}$_{(1)}$ & 3.10 {\scriptsize $\pm$ 0.04}$_{(5)}$ & 6.8\\
    
    HyperIMTS* (2025) & \textbf{4.59 {\scriptsize $\pm$ 0.03}}$_{(1)}$ &  4.50 {\scriptsize $\pm$ 0.04}$_{(11)}$  &  1.72 {\scriptsize $\pm$ 0.03}$_{(6)}$ & 7.23 {\scriptsize $\pm$ 0.04}$_{(4)}$ & 4.36 {\scriptsize $\pm$ 0.06}$_{(19)}$  & 4.12 {\scriptsize $\pm$ 0.08}$_{(14)}$ & 5.21 {\scriptsize $\pm$ 0.03}$_{(6)}$ & 3.21 {\scriptsize $\pm$ 0.03}$_{(7)}$ & 8.5\\ 
    \midrule
    \textbf{KAFNet (Ours)} &  5.88 {\scriptsize $\pm$ 0.01}$_{(5)}$ & \textbf{ 3.52 {\scriptsize $\pm$ 0.01}}$_{(1)}$ & \textbf{ 1.59 {\scriptsize $\pm$ 0.02 }}$_{(1)}$ & \textbf{ 6.83 {\scriptsize $\pm$ 0.08}}$_{(1)}$ &  \textbf{2.54 {\scriptsize $\pm$ 0.08}}$_{(1)}$ & \textbf{ 3.06 {\scriptsize $\pm$ 0.07 }}$_{(1)}$ &  \underline{4.98 {\scriptsize $\pm$ 0.02}}$_{(2)}$ & \textbf{ 2.99 {\scriptsize $\pm$ 0.01 }}$_{(1)}$ & \textbf{1.6}\\
    \bottomrule
  \end{tabular}}
  \caption{We report mean $\pm$ standard deviation over five random seeds for MSE and MAE. The best result is in \textbf{bold}, and the second-best is \underline{underlined}. Results from models marked with * are obtained from our own re‑implementation under a unified setting for fair comparison, others are collected from \citep{tPatchGNN}. The subscript denotes the rank; when two models achieve the same MSE or MAE, we rank them according to their standard deviations.}
  \label{tab:mainresult}
\end{table*}

\subsection{Experimental Setup}

\paragraph{Datasets.}
To empirical evaluate the model performance on IMTS forecasting, we employ four datasets from three distinct domains. These include two from healthcare, \textbf{PhysioNet} \citep{physionet} with 41 variates and \textbf{MIMIC} \citep{mimic} with 96 variates; one from biomechanics, \textbf{Human Activity} with 12 variates; and one from climate science, \textbf{USHCN} \citep{USHCN} with 5 variates. Adhering to the protocol established in tPatchGNN \citep{tPatchGNN}, we split each dataset into training, validation, and test subsets with a 60\%:20\%:20\% distribution. 

\paragraph{Baselines.}
To facilitate a comprehensive and rigorous comparison, we select a diverse suite of baseline models, which are categorized into four distinct methodological groups. The first group, \textbf{MTS Forecasting \footnote{Concat future-time queries with the encoder output and feed it into an MLP forecasting head.}}, comprises regular multivariate time series forecasting models: DLinear \citep{DLinear}, TimesNet \citep{Timesnet}, PatchTST \citep{PatchTST}, Crossformer \citep{Crossformer}, GraphWaveNet \citep{GraphWaveNet}, MTGNN \citep{MTGNN}, StemGNN \citep{Stemgnn}, CrossGNN \citep{Crossgnn}, and FourierGNN \citep{FourierGNN}. The second group consists of models designed for \textbf{IMTS Classification \footnote{Replace the classification head with an MLP forecasting head.}}, including GRU-D \citep{Che2018}, SeFT \citep{seft}, RainDrop \citep{RainDrop}, and Warpformer \citep{Warpformer}. For the \textbf{IMTS Interpolation \footnote{Swap interpolation targets for queries to enable extrapolation.}} category, we select mTAND \citep{mTAND}. The final group focuses on \textbf{IMTS Forecasting} methods: Latent ODEs \citep{Latent-ODE}, CRU \citep{CRU}, Neural Flows \citep{Neural-Flows}, tPatchGNN \citep{tPatchGNN}, GraFITi \citep{GraFITi}, TimeCHEAT \citep{TimeCHEAT}, and HyperIMTS \citep{HyperIMTS}.

\paragraph{Implementation Details.}\label{sec:implement}
All experiments are conducted on a single NVIDIA RTX A6000 GPU. During the training phase, models are optimized using the Adam \citep{kingma2017adam} optimizer to minimize the loss function as introduced in Eq. \eqref{eq:loss}. The predictive performance is evaluated using two standard metrics: Mean Squared Error (MSE) and Mean Absolute Error (MAE). 


\subsection{Main Results}
Table \ref{tab:mainresult} summarizes the results. We observe that KAFNet consistently outperforms the state‑of‑the‑art methods. Our improvements pass the Friedman test on dataset‑level ranks \(\alpha=0.05\), with Nemenyi post‑hoc confirming KAFNet’s pairwise superiority over the compared baselines. This superior performance stems from its bold adoption of CPA, which mitigates inter‑series asynchrony and unifies temporal resolution across variates. Moreover, KAFNet excels on high‑dimensional IMTS, e.g., the MIMIC dataset with 96 variates where pre‑aligned sequences can become prohibitively long.


\subsection{Hyper-parameter Analysis}
\begin{figure}[!ht]
  \centering
  \subfloat[Number of Kernels in TKA.]{%
    \includegraphics[width=0.49\linewidth]{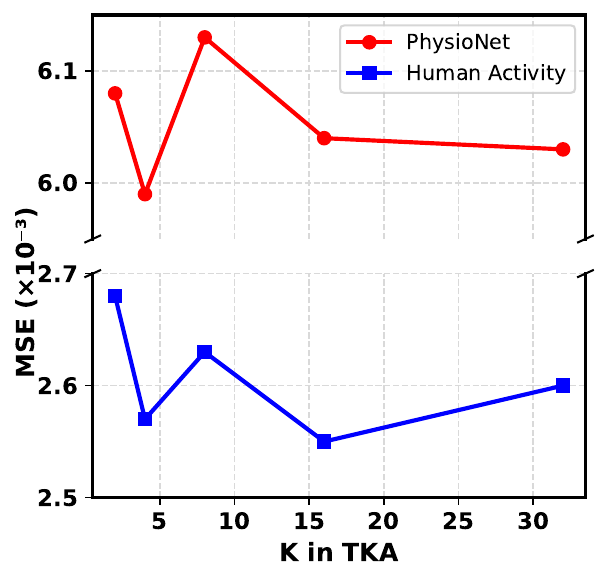}%
    \label{fig:hyper_k}%
  }\hfill
  \subfloat[Hidden State Dimension.]{%
    \includegraphics[width=0.49\linewidth]{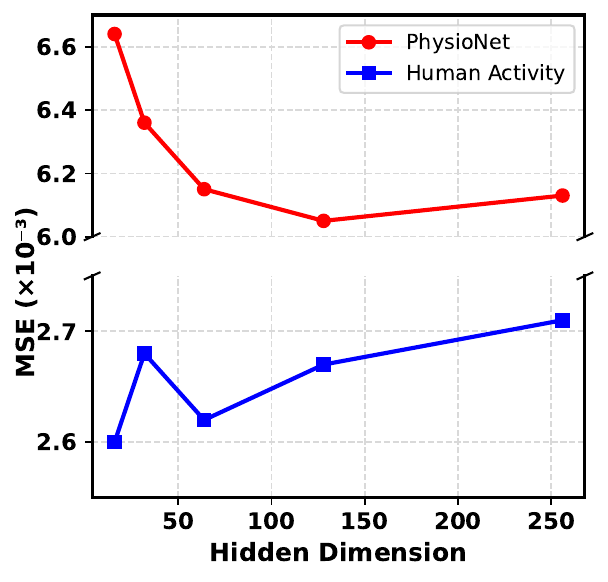}%
    \label{fig:hyper_Hiddendim}%
  }
  \caption{Sensitivity of MSE to (a) the number of Gaussian kernels in TKA and (b) the hidden state dimension in Eq. \eqref{eq:tka_proj}, FLA and the Output Layer, on two IMTS datasets.}
  \label{fig:hyperparam}
\end{figure}

We focus on two key hyper‑parameters in
KAFNet: (1) the number of Gaussian kernels in the TKA module; and (2) the hidden state dimension used in Eq. \eqref{eq:tka_proj}, FLA blocks and the final Output Layer. Fig. \ref{fig:hyperparam} illustrates how the MSE varies as each of these parameters is changed. Notably, adding more kernels to TKA does not always improve accuracy because too many kernels can overlap excessively along the normalized timeline. Similarly, increasing the hidden state dimension can boost capacity but may incur heavier computation and even degrade performance if over‑parameterized.

\begin{table*}[t]
  \centering
  \footnotesize
  \scalebox{0.85}{%
  \begin{tabular}{c|cc|cc|cc|cc}
    \toprule
    \textbf{Dataset}
        & \multicolumn{2}{c|}{PhysioNet}
        & \multicolumn{2}{c|}{MIMIC}
        & \multicolumn{2}{c|}{Human Activity}
        & \multicolumn{2}{c}{USHCN} \\ 
    \cmidrule(lr){1-2}\cmidrule(lr){2-4}\cmidrule(lr){4-5}\cmidrule(lr){5-8}\cmidrule(l){8-9}
    \addlinespace[2pt]
       \textbf{Metric} & MSE$\times10^{-3}$ & MAE$\times10^{-2}$
        & MSE$\times10^{-2}$ & MAE$\times10^{-2}$
        & MSE$\times10^{-3}$ & MAE$\times10^{-2}$
        & MSE$\times10^{-1}$ & MAE$\times10^{-1}$ \\
    \addlinespace[2pt]
    \midrule
    KAFNet                & \bm{$5.88 \pm 0.01$} & \bm{$3.52 \pm 0.01$}
                         & \bm{$1.59 \pm 0.02$} & \bm{$6.83 \pm 0.08$}
                         & \bm{$2.54 \pm 0.08$} & \bm{$3.06 \pm 0.07$}
                         & \bm{$4.98 \pm 0.02$} & \bm{$2.99 \pm 0.01$}\\
    \midrule
    w/o CPA              & $ 6.21 \pm 0.05$ & $3.88 \pm 0.04$
                         & $1.69 \pm 0.02$ & $6.98 \pm 0.05$
                         & $2.70 \pm 0.05$ & $3.20 \pm 0.03$
                         & $5.04 \pm 0.02$ & $3.10 \pm 0.02$ \\
    w/o Pre‑Conv         & $6.42 \pm 0.02$ & $3.70 \pm 0.04$
                         & $1.62 \pm 0.02$ & $6.91 \pm 0.03$
                         & $2.66 \pm 0.03$ & $3.17 \pm 0.04$
                         & $5.06 \pm 0.02$ & $3.05 \pm 0.03$\\
    w/o T-Norm           & $6.37 \pm 0.02$ & $3.83 \pm 0.08$
                         & $1.73 \pm 0.03$ & $7.50 \pm 0.10$
                         & $2.66 \pm 0.04$ & $3.18 \pm 0.04$
                         & $5.14 \pm 0.05$ & $3.08 \pm 0.05$\\
    w/o TKA              & $6.95 \pm 0.04$ & $3.98 \pm 0.08$
                         & $1.74 \pm 0.03$ & $7.29 \pm 0.10$
                         & $4.21 \pm 0.03$ & $4.15 \pm 0.04$
                         & $5.07 \pm 0.03$ & $3.14 \pm 0.05$\\
    w/o FLA              & $6.26 \pm 0.06$ & $3.78 \pm 0.08$
                         & $1.79 \pm 0.04$ & $7.75 \pm 0.09$
                         & $2.71 \pm 0.08$ & $3.12 \pm 0.12$
                         & $5.23 \pm 0.04$ & $3.15 \pm 0.06$\\
    w/o FLA \& w/ SA     & $6.08 \pm 0.02$ & $3.64 \pm 0.03$
                         & $1.67 \pm 0.05$ & $7.11 \pm 0.02$
                         & $2.57 \pm 0.06$ & $3.09 \pm 0.07$
                         & $5.20 \pm 0.06$ & $3.02 \pm 0.04$\\
    \bottomrule
  \end{tabular}}
  \caption{Ablation results on four datasets evaluated by MSE and MAE (mean $\pm$ standard). The best results are in \textbf{bold}.}
  \label{tab:ablation_results}
\end{table*}
 
\subsection{Ablation Study}

Given that KAFNet comprises several modular components for IMTS representation learning including CPA, the Pre‑Convolution module, TKA, and FLA blocks, we conduct an ablation study on four public IMTS datasets to assess each component’s contribution. As shown in Table \ref{tab:ablation_results}, removing any single module consistently degrades performance compared to the full KAFNet. Specifically, the removal of CPA leads to a significant decline in forecasting performance, thereby underscoring its essential role in IMTS modeling by alleviating inter-variate asynchrony. Moreover, two architecture‑agnostic designs: Pre‑Convolution and T‑Norm both yield notable improvements and could be adopted in other IMTS models. Furthermore, substituting our FLA with conventional softmax attention (SA) also results in inferior accuracy, underscoring the effectiveness of the proposed FLA in fostering the modeling of inter-variate correlations.

\begin{figure}[!ht]
  \centering
  \subfloat[Number of Parameters (K).]%
  {\includegraphics[width=0.49\linewidth]{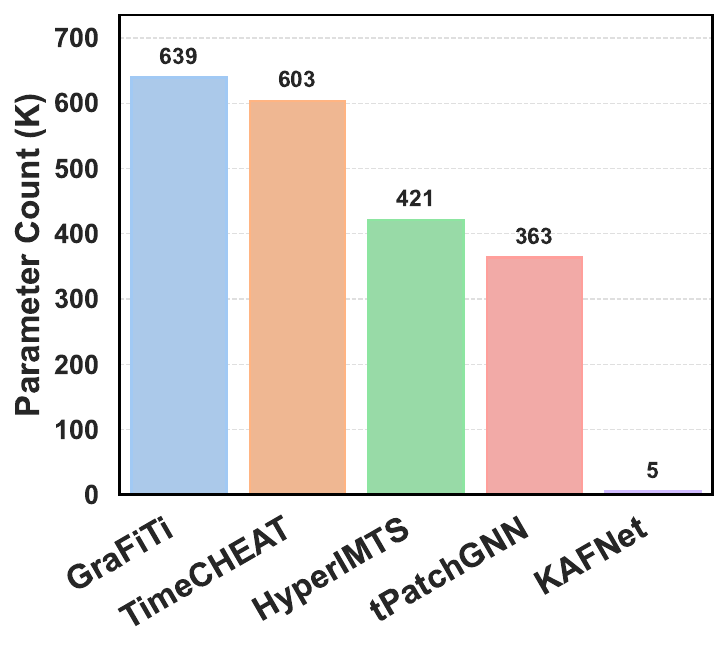}%
   \label{fig:num_parameters}}
  \hfill
  \subfloat[FLOPs (B).]%
  {\includegraphics[width=0.49\linewidth]{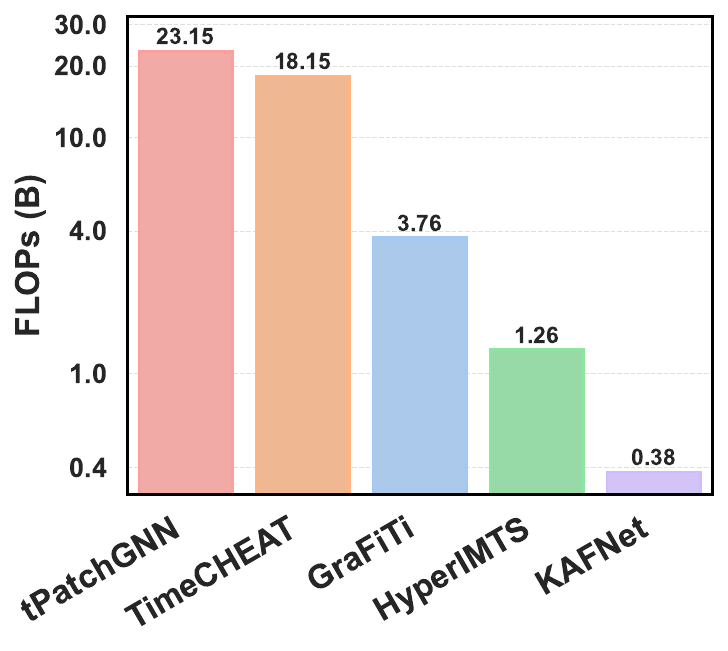}%
   \label{fig:flops}}
  \par\smallskip
  \subfloat[Training Time (s).]%
  {\includegraphics[width=0.49\linewidth]{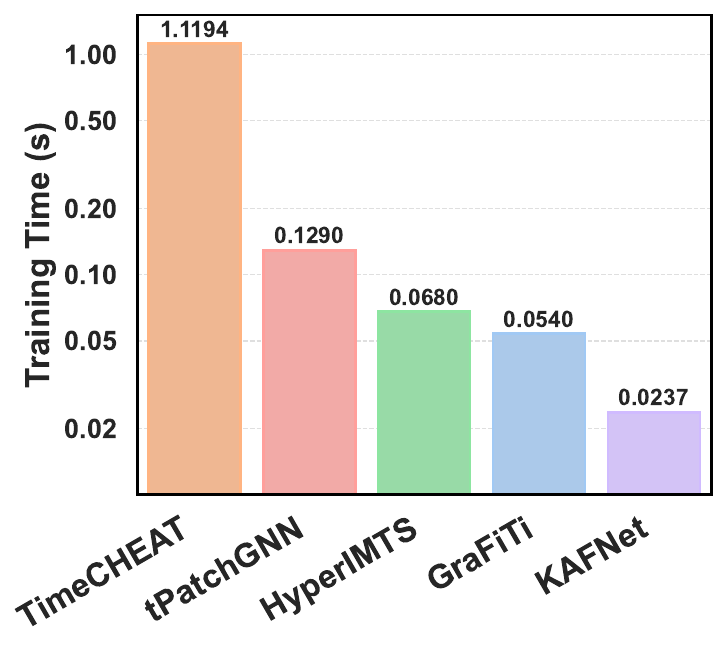}%
   \label{fig:training_time}}
  \hfill
  \subfloat[Inference Time (s).]%
  {\includegraphics[width=0.49\linewidth]{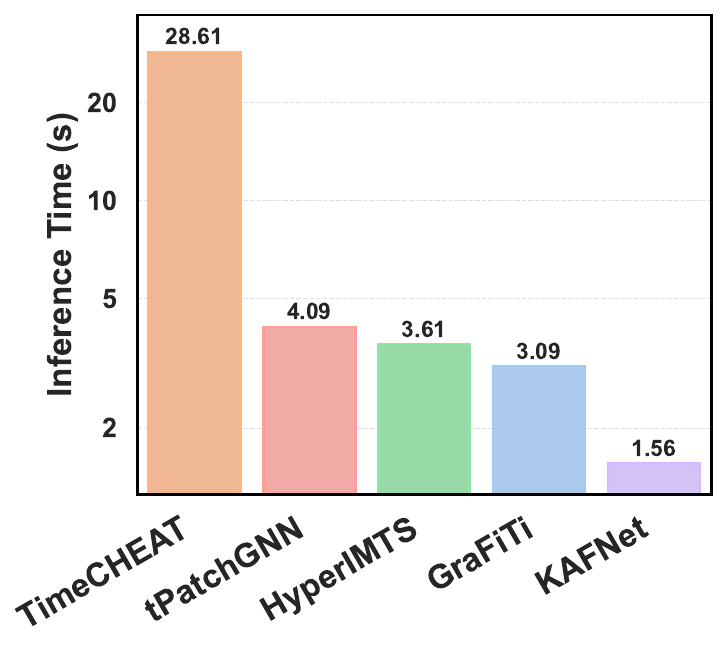}%
   \label{fig:inference_time}}
  \caption{Comparison of the number of parameters (K), FLOPs (B), average training time per batch per epoch (s), and total inference time (s) of KAFNet and four strong baselines for IMTS forecasting. All statistics are collected on the MIMIC dataset with a batch size of 32 to ensure a fair comparison. Lower values indicate higher efficiency.}
  \label{fig:model_performance_comparison}
\end{figure}

\subsection{Efficiency Analysis}
As shown in Fig. \ref{fig:radar}, KAFNet consistently outperforms strong lightweight baselines in efficiency. We therefore present a detailed analysis on the MIMIC dataset, evaluating efficiency across four dimensions: (i) number of trainable parameters, (ii) FLOPs, (iii) training time per batch per epoch, and (iv) inference time per batch. As shown in Fig. \ref{fig:model_performance_comparison}, KAFNet delivers a clear efficiency advantage over graph-based baselines. Panels (a) and (b) reveal that KAFNet attains the best predictive performance with only 5K parameters and 0.38B FLOPs (several orders of magnitude fewer than its counterparts), while panels (c) and (d) confirm that it also achieves the fastest training and inference.

\subsection{Comparison of FLA and SA}

\begin{figure}[!ht]
    \centering
    \includegraphics[width=\linewidth]{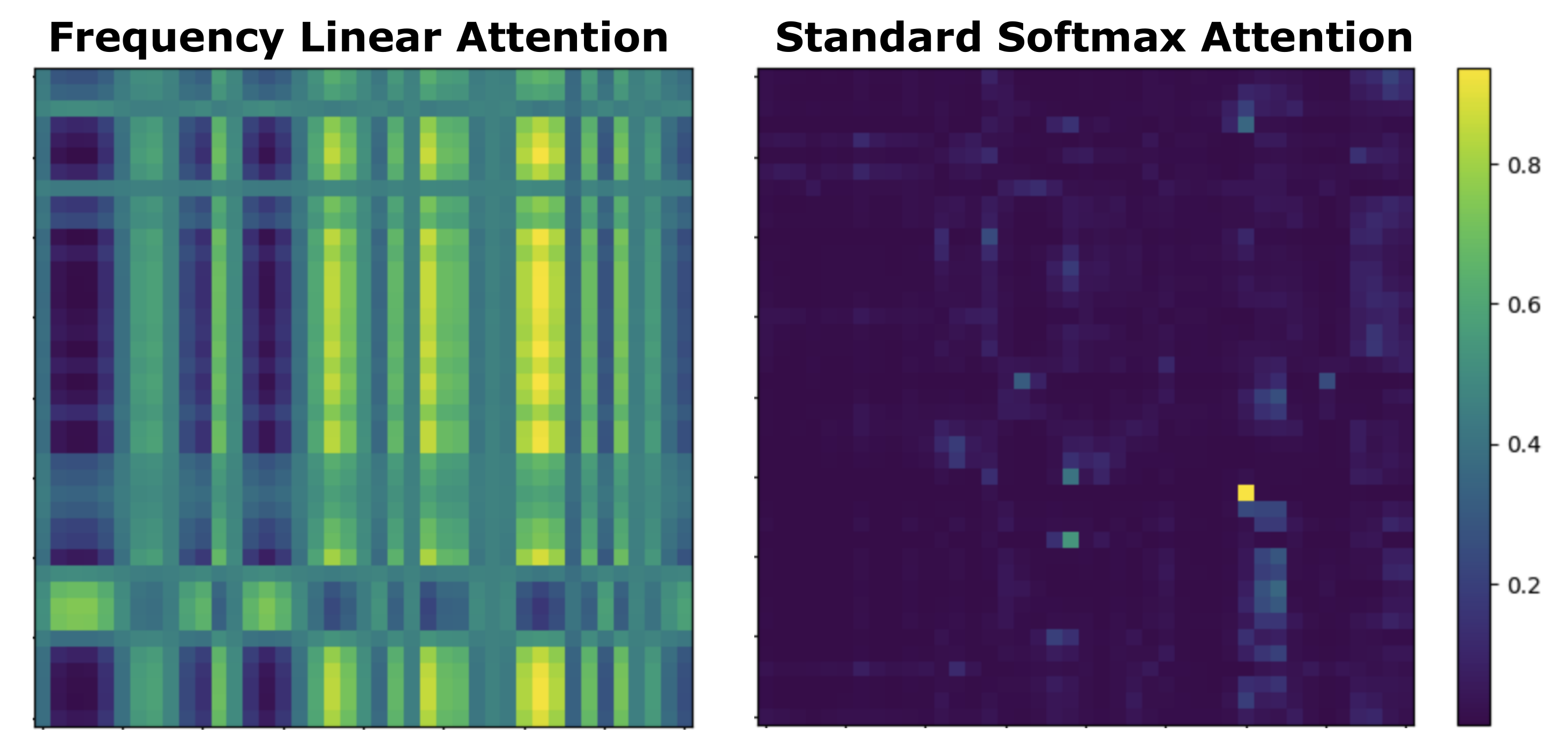}
    \caption{Visualization of attention maps collected from PhysioNet: the left panel shows maps from KAFNet with FLA, while the right panel shows maps from KAFNet in which FLA is replaced by standard Softmax Attention (SA).}
    \label{fig:attentionmap}
\end{figure}

To illustrate FLA’s superiority over the vanilla Softmax Attention (SA), Fig. \ref{fig:attentionmap} compares their attention maps. The distributions of attention scores produced by FLA and by conventional SA diverge markedly. The FLA map spans almost the entire color scale, indicating that each query variate assigns sharply different weights to different keys. In contrast, the SA map is dominated by values confined to a narrow, low‑magnitude band, with only a few isolated pixels exhibiting slightly higher scores. This broader dynamic range allows FLA to selectively amplify or suppress inter‑variate dependencies. Moreover, as Table \ref{tab:efficiency_fla_sa} shows, FLA also achieves explicit computational savings compared to SA.

\begin{table}[!ht]
    \centering
    \renewcommand{\multirowsetup}{\centering}
			\setlength{\tabcolsep}{2pt}
			\scalebox{0.83}{
			\begin{tabular}{c|c|c|c|c|c}
				\toprule
			 & Memory $\downarrow$ & Parameters  $\downarrow$ & FLOPs $\downarrow$& Train. Speed $\downarrow$& Infer. Speed $\downarrow$\\
			    \midrule
                  FLA & \textbf{890MB} &  \textbf{118.4K}  & \textbf{360.5M} & \textbf{ 0.023s} &  \textbf{1.23s} \\
                  SA & 1022MB &  125.4K & 378.2M &   0.039s  & 1.30s\\
				\bottomrule
			\end{tabular}}
        \caption{Efficiency Analysis on the PhysioNet dataset.}
    \label{tab:efficiency_fla_sa}
\end{table}

\section{Conclusion and Future Work}
This paper revisits Canonical Pre‑Alignment (CPA) for irregular multivariate time series (IMTS) forecasting and resolves the long‑standing tension between CPA’s alignment benefits and its computational overhead. We present \textbf{KAFNet}, a compact CPA‑based forecasting model that restores the advantages of pre‑aligned modeling while remaining highly efficient. Across representative IMTS forecasting benchmarks, KAFNet achieves strong predictive accuracy with fewer parameters and lower training/inference cost than leading graph‑based alternatives. However, our study is restricted to forecasting, and the evaluated datasets cover a limited set of domains. Future work will focus on extending the approach to other IMTS-related downstream tasks including classification, interpolation, and anomaly detection, and on scaling evaluations in transportation and energy IMTS scenarios under realistic deployment constraints.


\section*{Acknowledgment}
This work is mainly supported by the National Natural Science Foundation of China (No. 62402414). This work is also supported by the Guangdong Basic and Applied Basic Research Foundation (No. 2025A1515011994), the National Natural Science Foundation of China (No. 62406206), Guangzhou Municipal Science and Technology Project (No. 2023A03J0011), the Guangzhou Industrial Information and Intelligent Key Laboratory Project (No. 2024A03J0628), a grant from State Key Laboratory of Resources and Environmental Information System, and Guangdong Provincial Key Lab of Integrated Communication, Sensing and Computation for Ubiquitous Internet of Things (No. 2023B1212010007), and in part by the Research Grants Council of the Hong Kong Special
Administrative Region (Grant 16200021).

\bibliography{aaai2026}

@inproceedings{CSDI,
  author  = {Tashiro, Yusuke and Song, Jiaming and Song, Yang and Ermon, Stefano},
  title   = {CSDI: Conditional Score-based Diffusion Models for Probabilistic Time Series Imputation},
  booktitle = {Advances in Neural Information Processing Systems},
  year    = {2021}
}

@inproceedings{tPatchGNN,
  author    = {Zhang, Weijia and Yin, Chenlong and Liu, Hao and Zhou, Xiaofang and Xiong, Hui},
  title     = {Irregular Multivariate Time Series Forecasting: A Transformable Patching Graph Neural Networks Approach},
  booktitle = {International Conference on Machine Learning},
  year      = {2024}
}

@inproceedings{Neural-ODE,
  author    = {Chen, Tian Qi and Rubanova, Yulia and Bettencourt, Jesse and Duvenaud, David},
  title     = {Neural Ordinary Differential Equations},
  booktitle = {Advances in Neural Information Processing Systems},
  year      = {2018}
}

@inproceedings{Latent-ODE,
  author    = {Rubanova, Yulia and Chen, Tian Qi and Duvenaud, David},
  title     = {Latent Ordinary Differential Equations for Irregularly-Sampled Time Series},
  booktitle = {Advances in Neural Information Processing Systems},
  year      = {2019}
}

@inproceedings{CRU,
  author    = {Schirmer, Mona and Eltayeb, Mazin and Lessmann, Stefan and Rudolph, Maja},
  title     = {Modeling Irregular Time Series with Continuous Recurrent Units},
  booktitle = {International Conference on Machine Learning},
  year      = {2022}
}

@inproceedings{Neural-Flows,
  author    = {Bilos, Marin and Sommer, Johanna and Rangapuram, Syama Sundar and Januschowski, Tim and G{\"u}nnemann, Stephan},
  title     = {Neural Flows: Efficient Alternative to Neural ODEs},
  booktitle = {Advances in Neural Information Processing Systems},
  year      = {2021}
}

@inproceedings{HyperIMTS,
  author        = {Li, Boyuan and Luo, Yicheng and Liu, Zhen and Zheng, Junhao and Lv, Jianming and Ma, Qianli},
  title         = {HyperIMTS: Hypergraph Neural Network for Irregular Multivariate Time Series Forecasting},
  booktitle = {International Conference on Machine Learning},
  year      = {2025}
}

@inproceedings{Hi-Patch,
  author    = {Luo, Yicheng and Zhang, Bowen and Liu, Zhen and Ma, Qianli},
  title     = {Hi-Patch: Hierarchical Patch GNN for Irregular Multivariate Time Series},
  booktitle = {International Conference on Machine Learning},
  year      = {2025}
}

@inproceedings{MTGNN,
  author    = {Wu, Zonghan and Pan, Shirui and Long, Guodong and Jiang, Jing and Chang, Xiaojun and Zhang, Chengqi},
  title     = {Connecting the Dots: Multivariate Time Series Forecasting with Graph Neural Networks},
  booktitle = {ACM SIGKDD Conference on Knowledge Discovery and Data Mining},
  year      = {2020}
}

@inproceedings{comba,
      title={Comba: Improving Bilinear RNNs with Closed-loop Control}, 
      author={Jiaxi Hu and Yongqi Pan and Jusen Du and Disen Lan and Xiaqiang Tang and Qingsong Wen and Yuxuan Liang and Weigao Sun},
      booktitle ={Advances in Neural Information Processing Systems}, 
      year={2025},
}

@inproceedings{TimeCHEAT,
  author    = {Liu, Jiexi and Cao, Meng and Chen, Songcan},
  title     = {TimeCHEAT: A Channel Harmony Strategy for Irregularly Sampled Multivariate Time Series Analysis},
  booktitle = {AAAI Conference on Artificial Intelligence},
  year      = {2025}
}

@inproceedings{ASeer,
  title={Irregular traffic time series forecasting based on asynchronous spatio-temporal graph convolutional networks},
  author={Zhang, Weijia and Zhang, Le and Han, Jindong and Liu, Hao and Fu, Yanjie and Zhou, Jingbo and Mei, Yu and Xiong, Hui},
  booktitle={ACM SIGKDD Conference on Knowledge Discovery and Data Mining},
  year={2024}
}

@article{SPECTRUM,
  title={SPECTRUM: Spectral analysis of unevenly spaced paleoclimatic time series},
  author={Schulz, Michael and Stattegger, Karl},
  journal={Computers \& Geosciences},
  year={1997},
 
}

@article{Unleash,
  title={Unleash the power of pre-trained language models for irregularly sampled time series},
  author={Zhang, Weijia and Yin, Chenlong and Liu, Hao and Xiong, Hui},
  journal={arXiv preprint arXiv:2408.08328},
  year={2024}
}

@article{GenCast,
      title={Generalising Traffic Forecasting to Regions without Traffic Observations}, 
      author={Xinyu Su and Majid Sarvi and Feng Liu and Egemen Tanin and Jianzhong Qi},
      year={2025},
      journal={arXiv preprint arXiv:2508.08947}, 
}

@inproceedings{shen2023non,
  title={Non-autoregressive conditional diffusion models for time series prediction},
  author={Shen, Lifeng and Kwok, James},
  booktitle={International Conference on Machine Learning},
  year={2023},
}

@misc{MuSiCNet,
      title={MuSiCNet: A Gradual Coarse-to-Fine Framework for Irregularly Sampled Multivariate Time Series Analysis}, 
      author={Jiexi Liu and Meng Cao and Songcan Chen},
      howpublished  = {arXiv preprint arXiv:2412.01063},
      year = {2024}
}

@misc{wavets,
  author        = {Zhou, Ziyu and Hu, Jiaxi and Wen, Qingsong and Kwok, James T. and Liang, Yuxuan},
  title         = {Multi‑Order Wavelet Derivative Transform for Deep Time Series Forecasting},
  howpublished  = {arXiv preprint arXiv:2505.11781},
  year          = {2025}
}

@inproceedings{SDformer,
  author    = {Zhou, Ziyu and Lyu, Gengyu and Huang, Yiming and Wang, Zihao and Jia, Ziyu and Yang, Zhen},
  title     = {SDformer: Transformer with Spectral Filter and Dynamic Attention for Multivariate Time Series Long‑term Forecasting},
  booktitle = {International Joint Conference on Artificial Intelligence},
  year      = {2024}
}

@misc{TSFool,
      title={TSFool: Crafting Highly-Imperceptible Adversarial Time Series through Multi-Objective Attack}, 
      author={Yanyun Wang and Dehui Du and Haibo Hu and Zi Liang and Yuanhao Liu},
      howpublished  = {arXiv preprint arXiv:2209.06388},
      year = {2024}
}

@inproceedings{rff,
  title={Random features for large-scale kernel machines},
  author={Rahimi, Ali and Recht, Benjamin},
  booktitle={Advances in neural information processing systems},
  year={2007}
}

@misc{fang2025,
      title={Unraveling Spatio-Temporal Foundation Models via the Pipeline Lens: A Comprehensive Review}, 
      author={Yuchen Fang and Hao Miao and Yuxuan Liang and Liwei Deng and Yue Cui and Ximu Zeng and Yuyang Xia and Yan Zhao and Torben Bach Pedersen and Christian S. Jensen and Xiaofang Zhou and Kai Zheng},
      year={2025},
      howpublished={arXiv preprint arXiv:2506.01364}
}

@misc{ruan,
      title={Vision-Enhanced Time Series Forecasting via Latent Diffusion Models}, 
      author={Weilin Ruan and Siru Zhong and Haomin Wen and Yuxuan Liang},
      howpublished  ={arXiv preprint arXiv:2502.14887},
      year={2025}
}

@inproceedings{GraFITi,
  author    = {Yalavarthi, Vijaya Krishna and Madhusudhanan, Kiran and Scholz, Randolf and Ahmed, Nourhan and Burchert, Johannes and Jawed, Shayan and Born, Stefan and Schmidt‑Thieme, Lars},
  title     = {GraFITi: Graphs for Forecasting Irregularly Sampled Time Series},
  booktitle = {AAAI Conference on Artificial Intelligence},
  year      = {2024}
}

@inproceedings{shen2021time,
  title={Time series anomaly detection with multiresolution ensemble decoding},
  author={Shen, Lifeng and Yu, Zhongzhong and Ma, Qianli and Kwok, James T},
  booktitle={AAAI Conference on Artificial Intelligence},
  year={2021}
}

@inproceedings{PatchTST,
  author    = {Nie, Yuqi and Nguyen, Nam H. and Sinthong, Phanwadee and Kalagnanam, Jayant},
  title     = {A Time Series is Worth 64 Words: Long‑term Forecasting with Transformers},
  booktitle = {International Conference on Learning Representations},
  year      = {2023}
}

@inproceedings{Dualcast,
  title={Dualcast: A model to disentangle aperiodic events from traffic series},
  author={Su, Xinyu and Liu, Feng and Chang, Yanchuan and Tanin, Egemen and Sarvi, Majid and Qi, Jianzhong},
  year={2025},
  booktitle={International Joint Conference on Artificial Intelligence}
}

@inproceedings{GraphWaveNet,
  author    = {Wu, Zonghan and Pan, Shirui and Long, Guodong and Jiang, Jing and Zhang, Chengqi},
  title     = {Graph WaveNet for Deep Spatial‑Temporal Graph Modeling},
  booktitle = {International Joint Conference on Artificial Intelligence},
  year      = {2019}
}

@inproceedings{FourierGNN,
  author  = {Yi, Kun and Zhang, Qi and Fan, Wei and He, Hui and Hu, Liang and Wang, Pengyang and An, Ning and Cao, Longbing and Niu, Zhendong},
  title   = {FourierGNN: Rethinking Multivariate Time Series Forecasting from a Pure Graph Perspective},
  booktitle = {Advances in Neural Information Processing Systems},
  year    = {2023}
}

@misc{kingma2017adam,
  author        = {Kingma, Diederik P. and Ba, Jimmy},
  title         = {Adam: A Method for Stochastic Optimization},
  howpublished  = {arXiv preprint arXiv:1412.6980},
  year          = {2014}
}

@article{USHCN,
  author  = {Menne, M. J. and Williams, C. N. Jr. and Vose, R. S.},
  title   = {Long‑Term Daily and Monthly Climate Records from Stations Across the Contiguous United States (U.S. Historical Climatology Network)},
  journal = {United States Historical Climatology Network Dataset},
  year    = {2016}
}

@article{mimic,
  author  = {Johnson, Alistair E. W. and Pollard, Tom J. and Shen, Lu and Lehman, Li‑wei H. and Feng, Mengling and Ghassemi, Mohammad and Moody, Benjamin and Szolovits, Peter and Celi, Leo Anthony and Mark, Roger G.},
  title   = {MIMIC‑III: A Freely Accessible Critical Care Database},
  journal = {Scientific Data},
  year    = {2016}
}

@inproceedings{physionet,
  author    = {Silva, Ikaro and Moody, George and Scott, Daniel J. and Celi, Leo A. and Mark, Roger G.},
  title     = {Predicting In‑Hospital Mortality of ICU Patients: The PhysioNet/Computing in Cardiology Challenge 2012},
  booktitle = {Computing in Cardiology},
  year      = {2012}
}

@inproceedings{Che2018,
  author    = {Che, Zhengping and Purushotham, Sanjay and Cho, Kyunghyun and Sontag, David and Liu, Yan},
  title     = {Recurrent Neural Networks for Multivariate Time Series with Missing Values},
  booktitle = {Scientific Reports},
  year      = {2018}
}

@inproceedings{CirT,
  author    = {Liu, Yang and Zheng, Zinan and Cheng, Jiashun and Tsung, Fugee and Zhao, Deli and Rong, Yu and Li, Jia},
  title     = {CirT: Global Subseasonal‑to‑Seasonal Forecasting with Geometry‑inspired Transformer},
  booktitle = {International Conference on Learning Representations},
  year      = {2025}
}

@inproceedings{TimesURL, 
title={TimesURL: Self-Supervised Contrastive Learning for Universal Time Series Representation Learning},
author={Liu, Jiexi and Chen, Songcan},   booktitle = {AAAI Conference on Artificial Intelligence},
  year    = {2025} 
}

@inproceedings{AirRadar,
  author  = {Wang, Qiongyan and Xia, Yutong and Zhong, Siru and Li, Weichuang and Wu, Yuankai and Cheng, Shifen and Zhang, Junbo and Zheng, Yu and Liang, Yuxuan},
  title   = {AirRadar: Inferring Nationwide Air Quality in China with Deep Neural Networks},
  booktitle = {AAAI Conference on Artificial Intelligence},
  year    = {2025}
}

@inproceedings{ContiFormer,
  author    = {Chen, Yuqi and Ren, Kan and Wang, Yansen and Fang, Yuchen and Sun, Weiwei and Li, Dongsheng},
  title     = {ContiFormer: Continuous‑Time Transformer for Irregular Time Series Modeling},
  booktitle = {Advances in Neural Information Processing Systems},
  year      = {2023}
}

@misc{APN,
  author        = {Liu, Xvyuan and Qiu, Xiangfei and Wu, Xingjian and Li, Zhengyu and Guo, Chenjuan and Hu, Jilin and Yang, Bin},
  title         = {Rethinking Irregular Time Series Forecasting: A Simple yet Effective Baseline},
  howpublished  = {arXiv preprint arXiv:2505.11250},
  year          = {2025}
}

@inproceedings{Warpformer,
  author    = {Zhang, Jiawen and Zheng, Shun and Cao, Wei and Bian, Jiang and Li, Jia},
  title     = {Warpformer: A Multi‑scale Modeling Approach for Irregular Clinical Time Series},
  booktitle = {ACM SIGKDD Conference on Knowledge Discovery and Data Mining},
  year      = {2023}
}

@inproceedings{DLinear,
  author    = {Zeng, Ailing and Chen, Muxi and Zhang, Lei and Xu, Qiang},
  title     = {Are Transformers Effective for Time Series Forecasting?},
  booktitle = {AAAI Conference on Artificial Intelligence},
  year      = {2023}
}

@inproceedings{TimesNet,
  author    = {Wu, Haixu and Hu, Tengge and Liu, Yong and Zhou, Hang and Wang, Jianmin and Long, Mingsheng},
  title     = {TimesNet: Temporal 2D‑Variation Modeling for General Time Series Analysis},
  booktitle = {International Conference on Learning Representations},
  year      = {2023}
}

@inproceedings{Crossformer,
  author    = {Wang, Wenxiao and Yao, Lu and Chen, Long and Lin, Binbin and Cai, Deng and He, Xiaofei and Liu, Wei},
  title     = {CrossFormer: A Versatile Vision Transformer Hinging on Cross‑scale Attention},
  booktitle = {International Conference on Learning Representations},
  year      = {2022}
}

@inproceedings{StemGNN,
  author    = {Cao, Defu and Wang, Yujing and Duan, Juanyong and Zhang, Ce and Zhu, Xia and Huang, Congrui and Tong, Yunhai and Xu, Bixiong and Bai, Jing and Tong, Jie and Zhang, Qi},
  title     = {Spectral Temporal Graph Neural Network for Multivariate Time‑series Forecasting},
  booktitle = {Advances in Neural Information Processing Systems},
  year      = {2020}
}

@inproceedings{CrossGNN,
  author    = {Huang, Qihe and Shen, Lei and Zhang, Ruixin and Ding, Shouhong and Wang, Binwu and Zhou, Zhengyang and Wang, Yang},
  title     = {CrossGNN: Confronting Noisy Multivariate Time Series via Cross Interaction Refinement},
  booktitle = {Advances in Neural Information Processing Systems},
  year      = {2023}
}

@inproceedings{GRU-ODE,
  author    = {De Brouwer, Edward and Simm, Jaak and Arany, Adam and Moreau, Yves},
  title     = {GRU‑ODE‑Bayes: Continuous Modeling of Sporadically‑Observed Time Series},
  booktitle = {Advances in Neural Information Processing Systems},
  year      = {2019}
}

@inproceedings{SeFT,
  author    = {Horn, Max and Moor, Michael and Bock, Christian and Rieck, Bastian and Borgwardt, Karsten M.},
  title     = {Set Functions for Time Series},
  booktitle = {International Conference on Machine Learning},
  year      = {2020}
}

@inproceedings{mTAND,
  author    = {Shukla, Satya Narayan and Marlin, Benjamin M.},
  title     = {Multi‑Time Attention Networks for Irregularly Sampled Time Series},
  booktitle = {International Conference on Learning Representations},
  year      = {2021}
}

@inproceedings{RainDrop,
  author    = {Zhang, Xiang and Zeman, Marko and Tsiligkaridis, Theodoros and Zitnik, Marinka},
  title     = {Graph‑Guided Network for Irregularly Sampled Multivariate Time Series},
  booktitle = {International Conference on Learning Representations},
  year      = {2022}
}

\end{document}